%% file: main.tex
\documentclass[11pt]{article}
 
\usepackage[margin=1in]{geometry}

\usepackage[authoryear,round]{natbib}

\title{\makebox[\textwidth][c]{Efficient Learning and Symmetry Discovery under Exact Invariances}} 

\newcommand{\equalcontrib}{\textsuperscript{\ensuremath{\dagger}}} \newcommand{\affilone}{\textsuperscript{\ensuremath{\ddagger}}} \newcommand{\affiltwo}{\textsuperscript{\S}} \newcommand{\affilthree}{\textsuperscript{\P}} \author{ Ashkan Soleymani\textsuperscript{\ensuremath{\dagger},\ensuremath{\ddagger}} \and Behrooz Tahmasebi\textsuperscript{\ensuremath{\dagger},\S} \and Patrick Jaillet\affilone \and Stefanie Jegelka\affilthree }

\date{}

\usepackage{mathtools}
\usepackage[most]{tcolorbox} 
\usepackage{xcolor}

\usepackage{algorithm}
\usepackage{algorithmic}

\usepackage{amsthm,amsmath,amssymb}
 
\theoremstyle{plain}
\newtheorem{theorem}{Theorem}[section]
\newtheorem{proposition}[theorem]{Proposition}

\theoremstyle{definition}

\theoremstyle{remark}
\newtheorem{remark}[theorem]{Remark}

\definecolor{darkblue}{rgb}{0.0,0.0,0.65}
\definecolor{darkred}{rgb}{0.68,0.05,0.0}
\definecolor{darkgreen}{rgb}{0.0,0.29,0.29}
\definecolor{darkpurple}{rgb}{0.47,0.09,0.29}
 
\usepackage[backref=page]{hyperref}
\hypersetup{
   colorlinks = true,
   citecolor  = darkblue,
   linkcolor  = darkred,
   filecolor  = darkblue,
   urlcolor   = darkblue,
 }
 \usepackage[capitalize,noabbrev]{cleveref}

 \input{notation}

\begin{document}

 \maketitle

\begingroup
\makeatletter
\renewcommand{\@makefntext}[1]{%
  \setlength{\parindent}{1.5em}%
  \setlength{\parskip}{\smallskipamount}%
  #1%
}
\makeatother
\renewcommand{\thefootnote}{}
\footnotetext{%
\equalcontrib These authors contributed equally to this work.\par
\affilone MIT Laboratory for Information and Decision Systems (LIDS).
Emails: \texttt{\{ashkanso,jaillet\}@mit.edu}\par
\affiltwo Harvard John A. Paulson School of Engineering and Applied Sciences,
Harvard University, Cambridge, MA 02138, USA.
Email: \texttt{behrooz\_tahmasebi@seas.harvard.edu}\par
\affilthree Technical University of Munich (CIT, MCML, MDSI) and
MIT Computer Science and Artificial Intelligence Laboratory (CSAIL).
Email: \texttt{stefanie.jegelka@tum.de}%
}
\endgroup

\begin{abstract}
Learning with group invariances is central to many scientific and geometric learning problems, yet its computational foundations remain poorly understood. Even for classical supervised regression settings, it has been unclear whether one can efficiently compute a regression function that is \emph{exactly invariant} to a given group action.
Recent work~\citep{soleymanilearning} showed that exact invariance can be enforced in polynomial time when the underlying group is finite and known, but left open the cases of infinite groups and unknown symmetries. In this paper, we resolve both challenges. First, we present the first polynomial-time algorithm for learning with exact group invariances that applies uniformly to finite and infinite groups. The runtime is polynomial in the data dimension and sample size, and independent of the group, while achieving strong generalization guarantees. This provides a computational explanation for the empirical success of invariant and equivariant methods in geometric machine learning and partially answers a recent open question in the literature~\citep{diaz2025invariant}. Second, we study learning in the \emph{symmetry discovery} setting, where the invariance group is unknown. Focusing on the subgroup lattice of a finite group, we show that exact symmetries can be identified from data and exploited for learning in polynomial time. For regression over finite-dimensional feature spaces, our algorithm provably recovers the underlying symmetry, matches the minimax-optimal sample complexity of the known-symmetry setting, and runs in time polynomial in the data dimension and sample size. Our analysis relies on tools from random Cayley graphs and expander theory, which may be of independent interest.
\end{abstract}

\vspace{0.5em} \noindent\textbf{Keywords:} group invariances, symmetry, learning, computational statistics \vspace{0.5em}

 \clearpage

 \tableofcontents
 
\section{Introduction}
\label{sec:intro}

Invariances have been central to learning theory since the earliest days of machine learning, and their importance has only deepened with modern advances~\citep{hinton1987learning,kondor2008group}. Models that explicitly respect underlying symmetries consistently deliver remarkable gains in practice, combining efficiency with strong generalization~\citep{bronstein2017geometric, weber2025geometric}. This empirical success has fueled an active line of research, though much of the theory remains focused on classic questions of expressivity, sample complexity, and testing~\citep{elesedy2021provably,bietti2021sample,behboodi2022pac,tahmasebi2023exact,mei2021learning,kiani2024on,soleymani2025robust,chen2023sample,tahmasebi2025generalization,diaz2025invariant,petrache2023approximation, tahmasebi24a}. What is still missing is a systematic understanding of the computational price of incorporating invariances, even in fundamental settings such as the classical formulation of supervised regression. 

How do we actually build invariances into algorithms? The most direct answer is brute force: expand the dataset through \emph{augmentation} or sum over symmetries via \emph{group averaging}. Unfortunately, both become infeasible once the symmetry group is large, sometimes even growing super-exponentially with input dimension. More ``clever'' alternatives, like \emph{canonicalization} or \emph{frame averaging}, often replace intractability with discontinuities, poor scalability, or the need for group-specific designs~\citep{dym2024equivariant,tahmasebi2025regularity}.

\subsection{Our Contributions}

In this paper, we advance the foundational understanding of learning under exact group invariances by establishing the computational tractability of invariant learning beyond the finite-group regime and by developing provably efficient methods for symmetry discovery from data.

\paragraph{Settling the computational complexity for infinite groups.}

Given the potentially prohibitive size of the symmetry group, one might expect learning with exact invariances to be computationally intractable, even in classical settings such as kernel regression. Surprisingly, recent work~\citep{soleymanilearning} introduced \emph{spectral averaging}, an exactly invariant algorithm that achieves favorable population risk with runtime $\mathrm{poly}(n, d, \log |G|)$, where $n$ is the number of samples, $d$ is the input dimension, and $|G|$ is the group cardinality. For \emph{finite groups}, this yields an effectively polynomial-time procedure, since $\log |G|$ is typically polynomial in $d$.

More precisely, spectral averaging decomposes functions into eigenfunction components, estimates the corresponding coefficients from data, and finally applies a projection onto the invariant subspace. Crucially, the projection operator is constructed using only $\mathcal{O}(\log |G|)$ group elements, leveraging the existence of small generating sets for finite groups.

This result naturally raises the question of whether similar guarantees can be achieved for \emph{infinite groups}. In particular, one would hope for algorithms whose runtime is independent of the group cardinality. However, a key ingredient in the analysis of~\citet{soleymanilearning} is that any finite group admits a generating set of size $\mathcal{O}(\log |G|)$, a property that fails dramatically for infinite groups, which may admit arbitrarily large generating sets or no finite generating set at all. As a result, existing techniques do not extend to this setting.

Our first contribution is to resolve this gap by proposing a polynomial-time algorithm for learning with exact group invariances that applies uniformly to \emph{both finite and infinite groups}. The runtime of our algorithm is \emph{independent of the group size}, depending only polynomially on the sample size $n$ and the input dimension $d$. In this sense, we provide an affirmative answer to the question of whether exact invariance can be enforced efficiently in the infinite-group regime. Our algorithm consists of a randomized preprocessing step that selects a specific small subset $S$ of the group, followed by a deterministic polynomial-time procedure. Despite its group-independent runtime, it achieves generalization guarantees comparable to those of spectral averaging.

\begin{theorem}[Informal version of \Cref{thm}]
Consider a supervised regression problem in a kernel setting with $n$ i.i.d.\ labeled samples from a $d$-dimensional domain, where the target function is invariant under a known group $G$, which may be finite or infinite. 
Then, there exists a randomized algorithm with runtime $\mathrm{poly}(n,d)$ such that, with high probability, it returns a $G$-invariant estimator with provable excess population risk guarantees.
The algorithm uses randomness only in a data-independent preprocessing step, where it samples a small subset $S\subseteq G$; conditioned on $S$, the procedure is deterministic.
\end{theorem}

The existence of such an algorithm provides a computational explanation for the empirical success of invariant and equivariant methods in geometric machine learning. Moreover, a recent open question in the literature~\citep{diaz2025invariant} concerns the design of computationally efficient algorithms for finite-dimensional invariant regression over general groups. While prior work proposes scalable solutions for specific group families, the existence of a polynomial-time algorithm applicable to \emph{arbitrary} groups has remained open. This work resolves that question in the affirmative.

\paragraph{Symmetry discovery for learning under invariances.}
While group invariances arise naturally in many applications, in a range of recent problems, particularly in scientific discovery and the identification of governing laws from data, the underlying symmetry group is not known a priori and must itself be inferred from data. This setting gives rise to the problem of \emph{symmetry discovery}.

In this setting, the learner must simultaneously identify the symmetry group and enforce the corresponding invariance constraints when learning the target function. This task is fundamentally more challenging than learning under a known symmetry, as it requires searching over a potentially large class of candidate groups.  Since finite groups may contain exponentially many subgroups, naive approaches based on enumerating candidate subgroups can be computationally infeasible.

To advance this direction, we study symmetry discovery for learning under exact invariances within the \emph{subgroup lattice} of a given finite group, under a bounded-index assumption. Specifically, we consider a known finite reference group $G$ and assume that the true symmetry group $H \subseteq G$ is an unknown subgroup governing the invariances of the task. Our goal is to recover $H$ by returning a generating set for it and to learn a regression function that is exactly invariant with respect to $H$ while achieving strong generalization guarantees.

Surprisingly, we show that exact symmetries can be identified from data and exploited for learning in polynomial time. For supervised regression problems with finite-dimensional kernels, we prove a stronger result: our algorithm provably recovers the underlying symmetry group, achieves the same minimax-optimal sample complexity as in the known-symmetry setting, and runs in time polynomial in the sample size and data dimension.

\begin{theorem}[Informal version of \Cref{thm:disc}]
Consider a supervised regression problem in a kernel setting with $n$ i.i.d.\ labeled samples from a $d$-dimensional domain, where the target function is invariant under an unknown group $H$. Assume that $H$ is a subgroup of a known finite group $G$ and satisfies $|H| \ge \kappa|G|$. Then, there exists a randomized algorithm with runtime $\mathrm{poly}\Big(n,d, \tfrac{1}{\kappa},\log |G|\Big)$ such that, with high probability, it returns a generating set of size $\mathcal{O}(\log |H|)$ for $H$ and an exactly $H$-invariant estimator with provable excess population risk guarantees. The algorithm uses randomness only in a data-independent preprocessing step, where it samples a small subset $S\subseteq G$; conditioned on $S$, the procedure is deterministic.
\end{theorem}

\paragraph{Tools and ideas.}
Our analysis relies on tools from the theory of random Cayley graphs and expanders~\citep{alon1994random}, which we believe may be of independent interest for the theoretical study of learning under group invariances.

For settling the computational complexity of learning under exact invariances for infinite groups, our approach is conceptually simple: we sample a small number of group elements uniformly at random and show that, with high probability, these samples suffice to construct an accurate projection onto the space of invariant functions. This idea is formalized in Algorithm~\ref{alg}.

For the symmetry discovery setting, we again leverage random sampling over the reference group $G$. We design a rejection-sampling–type procedure that probabilistically “hits” the unknown subgroup $H$, enabling the recovery of a generating set of size $\mathcal{O}(\log |H|)$. The correctness and efficiency of this procedure follow from expansion properties of random Cayley graphs, which allow us to control both the runtime and the probability of success.

\section{Related Work}

Invariance has long been recognized as a powerful inductive bias in statistical learning. Incorporating symmetry into models can reduce sample complexity and improve generalization by restricting hypotheses to symmetry-respecting subsets~\citep{hinton1987learning,poggio1992recognition,haussler1999convolution,kondor2008group,sokolic2017generalization}. A classical way of achieving this is through invariant kernels, constructed by \emph{group averaging}~\citep {scholkopf2002learning,haasdonk2007invariant}. Beyond this idea, alternative strategies include \emph{frame averaging} \citep{puny2021frame}, \emph{canonicalization} \citep{kaba2023equivariance,ma2024canonization}, \emph{random projections} \citep{dym2024low}, and \emph{parameter sharing} \citep{ravanbakhsh2017equivariance}. Each of these methods has distinct strengths, but also drawbacks. In particular, canonicalization and frame averaging can introduce discontinuities or violate smoothness assumptions, a limitation highlighted in recent work on equivariant frames \citep{dym2024equivariant}. 

Beyond kernel methods, symmetries have played a central role in the design of specialized learning architectures. Graph Neural Networks (GNNs) exploit permutation symmetries in graphs \citep{scarselli2008graph,xu2018powerful}, Convolutional Neural Networks (CNNs) leverage translation invariance in image data \citep{krizhevsky2012imagenet,li2021survey}, and PointNet architectures encode permutation invariance for point clouds \citep{qi2017pointnet,qi2017pointnetplus}. Symmetry principles have also been integrated into generative models, including permutation-invariant normalizing flows and equivariant flows \citep{bilovs2021scalable,niu2020permutation,kohler2020equivariant}. For a broad discussion on geometric invariances across modalities, we refer to the survey of \citet{bronstein2017geometric}.

Compared with these approaches, our contribution can be seen as a \emph{post-hoc invariantization} procedure: starting from (spectral) estimates of regression coefficients, we project onto the fixed-point subspaces determined by a (randomized) set of group elements. This yields estimators that are \emph{exactly} invariant, with statistical guarantees matching standard bounds. Prior work on finite groups established this using generating sets of size at most $\log|G|$ \citep{soleymanilearning}, while our randomized subset selection algorithm removes the dependence on $|G|$ altogether, extending polynomial-time learning with exact invariances to infinite groups \citep{soleymanilearning}.

Learning under unknown (but existing) symmetries, often referred to as \emph{symmetry discovery}, has recently attracted significant attention. A wide range of methods have been proposed, primarily based on deep learning architectures and Lie-algebraic techniques, including approaches for discovering continuous symmetries \citep{desai2022symmetry, yang2023generative, JMLR:v26:24-2175, dehmamy2021automatic, romero2022learning, ko2024learning, hu2025symmetry} as well as methods tailored to finite groups \citep{huh2025a}. Related work explores symmetry identification via gradients of neural network layers \citep{van2023learning}, flow-matching techniques for Lie group symmetries \citep{park2025discovering}, or joint procedures that simultaneously discover and enforce symmetries in learned models \citep{otto2025unified}, or enforce them approximately \citep{tahmasebi2025achieving}. While these approaches demonstrate strong empirical performance, they typically lack rigorous statistical or computational guarantees.

Symmetry discovery has also been studied beyond the data domain, for example in latent representations learned by neural networks \citep{yang2024latent}. In this setting, discovered symmetries may extend beyond affine transformations and relate to underlying manifold structure \citep{shaw2024symmetry, bhat2025atlasd}. Other approaches rely on data augmentation strategies \citep{santos2025learning} or heuristic procedures \citep{karjol2025learning}. We note that symmetry discovery is distinct from the problem of hypothesis testing for the presence of a \emph{given} symmetry in data, which has been studied separately \citep{soleymani2025robust}.

Symmetry discovery is also relevant to dynamical systems arising in scientific applications \citep{tahmasebi2026adaptive}. For instance, \citet{calvo2025learning, calvo2025machine} propose methods for identifying finite group invariances in dynamical systems, while approaches for discovering infinitesimal Lie group generators have been developed in \citep{pmlr-v267-hu25o}. Additional related work studies symmetry discovery in latent or continuous-time dynamical models \citep{li2025latent, shaw2025cont, gabel2024data}. For further connections between symmetry discovery and differential equations or partial differential equations, see \citep{kreider2025model, hu2025governing, yang2025discovering, yang2024symmetry}.

\section{Problem Statement}

We begin by formalizing the learning setting and introducing the problem studied in this paper.

\paragraph{Data generation and function space.}
We consider a well-specified supervised learning problem on a smooth, compact,
boundaryless Riemannian manifold $\mathcal M$ of dimension $d$. Given $n$
independent samples $\mathcal S=\{(x_i,y_i)\}_{i=1}^n$, the inputs
$x_i\in\mathcal M$ are drawn uniformly (i.e., with respect to the canonical Riemannian
volume measure), and the labels are generated as
\[
    y_i=f^\star(x_i)+\epsilon_i,
\]
where $f^\star\in L^2(\mathcal M)$ is an unknown continuous regression function
and the noise variables $\epsilon_i$ are independent, zero mean, and have
variance at most $\sigma^2$. The uniformity assumption is made for simplicity; our
results extend to distributions with bounded densities.

The performance of an estimator $\widehat f$ is measured by its population risk
\[
    \mathcal R(\widehat f)
    :=\mathbb E\!\left[\|\widehat f-f^\star\|_{L^2(\mathcal M)}^2\right],
\]
where the expectation is over the randomness of the data.

\paragraph{Group actions and symmetries.}
Suppose that a known compact group $G$ acts smoothly and isometrically on $\mathcal M$, with the action denoted by $gx$ for any $g\in G$ and any $x\in\mathcal M$. A function $f^\star$ is said to be $G$-invariant if
\[
f^\star(gx)=f^\star(x)\qquad\text{for all } g\in G,\ x\in\mathcal M.
\]
In this setting, the goal is to construct an estimator $\widehat f$ that is both accurate and exactly invariant under the action of $G$.

Throughout the paper, we work with a Hilbert feature space
$\mathcal H\subseteq L^2(\mathcal M)$ that is closed under the group action, and
we assume that this action is isometric (unitary) on $\mathcal H$. This is a natural
compatibility condition between the feature space and the transformations: in
finite dimensions, it can be enforced by choosing an invariant inner product,
while in geometric settings it is canonically satisfied by many spectral
constructions. In particular, on manifold domains, Laplace--Beltrami eigenspaces
and their finite-dimensional truncations provide natural feature spaces that are
compatible with both the geometry of $\mathcal M$ and Sobolev regularity. For
example, when $\mathcal M$ is the sphere, these eigenspaces are the spaces of
spherical harmonics. We denote by $\mathcal H^G$ the subspace of $G$-invariant functions in $\mathcal H$.

We consider two representative regimes. First, in the finite-dimensional regime,
$\mathcal H$ is any finite-dimensional Hilbert feature space for which the
kernel, or equivalently the associated feature map, can be evaluated, for
instance through oracle access. In this setting, Empirical Risk Minimization (ERM) over
$\mathcal H$ attains the minimax-optimal risk
$\mathcal O(\dim(\mathcal H)/n)$. Second, in the Sobolev regime, we take
$\mathcal H=\mathcal H^s(\mathcal M)$ with $s>d/2$. In this case, Kernel Ridge
Regression (KRR) achieves the minimax-optimal risk rate
$\mathcal O(n^{-s/(s+d/2)})$, albeit with a naive computational cost of
$\mathcal O(n^3)$ when implemented using kernel evaluations
\citep{bach2024learning}. These two regimes are chosen for clarity and cover the
main examples of interest; the framework itself applies more broadly to kernels
and feature spaces compatible with the group action.

Note that even when $f^\star$ is $G$-invariant, standard learning procedures such as KRR with Sobolev kernels or ERM do not, in general, return invariant estimators \citep{soleymanilearning}. As a result, additional structure must be imposed to enforce invariance. A classical approach is \emph{group averaging}, either by averaging the kernel or the learned estimator over the group $G$. However, such methods require $\mathcal O(|G|)$ operations even to evaluate the estimator, rendering them computationally intractable for large or infinite groups.

A recent breakthrough by \citet{soleymanilearning} introduced \emph{spectral averaging}, which achieves the same minimax-optimal risk rate (that is $\mathcal O\!\left(n^{-s/(s+d/2)}\right)$ in Sobolev spaces and $\mathcal O(\dim(\mathcal H)/n)$ in finite-dimensional settings) while running in $\operatorname{poly}(n,d,\log |G|)$ time. This yields an exponential improvement over naive group averaging for finite groups. However, the dependence on $\log|G|$ fundamentally limits the applicability of spectral averaging to infinite groups, such as continuous rotation groups.

Consequently, any polynomial-time algorithm for learning with exact invariances under infinite groups must have a runtime that is independent of the group cardinality. The central goal of this paper is to address this challenge. In the following sections, we provide an affirmative answer by developing a novel algorithmic framework that enforces exact invariance while maintaining polynomial-time complexity independent of $|G|$.

\paragraph{Compactness and sampling from the group}
Our assumption that $G$ is compact should be understood as an assumption on the
effective transformation group acting on the data domain. Since our results only
depend on the induced action on $\mathcal M$, one may always quotient out the
kernel of the action and work with the corresponding faithful action. In this
paper, we assume that this effective group is a compact group acting smoothly on
the compact Riemannian manifold $\mathcal M$. In particular, this covers the
standard setting of compact Lie group actions, for which there is a unique Haar
probability measure. This measure provides the canonical notion of uniform
sampling from the group $G$.

We emphasize, however, that sampling from $G$ can be a separate algorithmic
problem, depending on how the group is represented. For example, if
$\mathcal M=S^1$ (the unit circle in two dimensions) and $G$ is a finite subgroup of equally spaced rotations, then
even specifying a uniformly random element requires $\Theta(\log |G|)$ random
bits. More generally, for groups specified implicitly, such as automorphism
groups of graphs, uniform generation can itself be computationally nontrivial.
We therefore separate this issue from the statistical and approximation
questions studied here, and assume oracle access to independent Haar-uniform
samples from $G$. This assumption is standard in many settings of interest and
is often mild; for instance, for several explicit matrix groups, finite cyclic
groups, and permutation groups given by suitable generators, uniform or nearly
uniform sampling can be implemented efficiently. For example, for permutation groups given by generators, this oracle can be implemented in polynomial time using standard Schreier--Sims methods, which also provide polynomial-time membership testing.

\paragraph{Unknown invariances and symmetry discovery.}
In many applications arising in scientific discovery, the symmetry group respected by the target function $f^\star\in\mathcal H$ is not known a priori and must be inferred from data. To model this setting, we consider a known finite reference group $G$ and assume that the true invariance group is an unknown subgroup $H\subseteq G$. The collection of all subgroups of $G$ forms a \emph{subgroup lattice} under set inclusion.

The goal of the symmetry discovery problem is to use a labeled dataset $\mathcal S$ to identify the underlying symmetry structure and to construct an estimator that is exactly invariant with respect to $H$, while achieving provable generalization guarantees and efficient runtime. Note that a naive brute-force search over all subgroups of $G$ is computationally infeasible, even when $|G|$ is only moderately large, due to the combinatorial size of the subgroup lattice.

To make the problem tractable, we focus on symmetry discovery within the class of \emph{bounded-index} subgroups. Specifically, we assume that the unknown subgroup $H$ satisfies
\[
[G:H]\coloneqq \frac{|G|}{|H|}\le \frac{1}{\kappa} \Longleftrightarrow |H| \ge \kappa |G|,
\]
for some parameter $\kappa \in [0,1]$. The index bound controls the size of the search space within the subgroup lattice: smaller values of $\kappa$ correspond to broader exploration over candidate subgroups, while larger values restrict attention to subgroups closer to $G$ itself.

\section{Main Results}\label{sec:mr}

In this section, we present our main results. We begin with the setting of learning under \emph{exact} invariances, where the symmetry group is known, and then turn to the more challenging problem of \emph{symmetry discovery}, in which the invariances must be inferred from data.

\subsection{Learning with Exact Invariances in Polynomial Time}

We start by considering the case where the underlying symmetry group is known and acts through a given representation. A key algorithmic component underlying our results is a randomized subset selection procedure, which adaptively constructs a small subset of group elements sufficient to characterize the invariant subspace.

To state our theoretical guarantees, we take a closer look at the spectral averaging framework of \citet{soleymanilearning}. Throughout, we consider hypothesis classes $\mathcal{H}$ that are either finite-dimensional, or Sobolev spaces of order $s > \tfrac{d}{2}$.

The spectral-averaging algorithm of \citet{soleymanilearning} shows that, for Sobolev spaces, the problem of learning with exact invariances can be reduced to solving an \emph{infinite collection of finite-dimensional convex quadratic programs with linear constraints}. Each program corresponds to an eigenspace of the Laplace--Beltrami operator associated with the data manifold $\mathcal{M}$, and the reduction relies on tools from differential geometry and spectral theory.

To obtain a computationally efficient procedure, \citet{soleymanilearning} truncate this infinite collection and solve only finitely many quadratic programs. This yields an approximation to the original optimization problem while incurring a negligible approximation error. Concretely, with a slight abuse of notation, let $\mathcal{H}$ denote the subspace of the Sobolev space $\mathcal{H}^s(\mathcal{M})$ spanned by the first
$r \coloneqq n^{1/(1+\alpha)} = \mathcal{O}(\sqrt{n})$ eigenfunctions, $\alpha \coloneqq \tfrac{2s}{d}$. 
This truncation is canonical and reduces learning in $\mathcal{H}^s(\mathcal{M})$ to a finite-dimensional problem in $\mathcal{H} \subseteq L^2(\mathcal{M})$, with approximation error that vanishes at a polynomial rate in $n$. Consequently,  we can focus on finite-dimensional hypothesis spaces with
$r \coloneqq \dim(\mathcal{H})$.

Fix an orthonormal basis $\{\phi_\ell\}_{\ell=1}^r$ of $\mathcal{H}$. Any function $f \in \mathcal{H}$ can then be identified with its coefficient vector
$f \in \mathbb{R}^r$. Let $D(g) \in \mathbb{R}^{r \times r}$ denote the matrix representation of the action of $g \in G$ on $\mathcal{H}$, defined by
$
(g  f)(x) \coloneqq f(g^{-1}x), \forall f \in \mathcal{H}.
$
Under this identification, enforcing invariance of $f$ under the group action amounts to the linear constraints $D(g) f = f$ for all $g \in G$.

The spectral-averaging framework, therefore, leads to the following optimization problem:
\[
\begin{aligned}
\min_{f \in \mathbb{R}^r} \quad
& \sum_{\ell=1}^{r} (f_{\ell} - \widetilde{f}_{\ell})^2 
\quad\text{s.t.} \quad
& D(g) f = f, \quad \forall g \in S,\quad \widetilde{f}_{\ell} \coloneqq \frac{1}{n} \sum_{i=1}^n y_i \phi_{\ell}(x_i),
\quad \forall \ell \in [r],
\end{aligned}
\]
and $S \subseteq G$ is a \emph{generating set} of the finite group $G$, meaning that every element of $G$ can be expressed as a finite word over $S$.

This deterministic procedure yields an estimator achieving population risk
$\mathcal{O}\!\left(n^{-s/(s+d/2)}\right)$ for Sobolev spaces, or
$\mathcal{O}(r/n)$ in finite-dimensional settings, and can be implemented in  
$\operatorname{poly}(n,d,\log |G|)$ time for finite groups.

The $\operatorname{poly}(\log |G|)$ dependence on the group size arises from the number of linear constraints imposed by the invariance conditions. Specifically, one constraint is required for each generator in $S$. Since any finite group admits a generating set of size at most $\log_2 |G|$, this accounts for the polylogarithmic dependence of the computational complexity on $|G|$. In contrast, for infinite groups, the cardinality $|G|$ is unbounded, and thus such a method fails to reduce the number of constraints.

In \Cref{alg}, we introduce a new randomized procedure that, with high probability, identifies a subset $S \subseteq G$ imposing a sufficient collection of linear constraints in the quadratic programs associated with each eigenspace of the spectral-averaging framework. Crucially, this procedure applies to both finite and infinite groups.

Specifically, the algorithm returns a subset $S$ of size
$\mathcal{O}\!\left(r^2 \log \tfrac{1}{\delta}\right)$
with probability at least $1-\delta$, as formalized in Proposition~\ref{prop:algo}. The runtime and the size of the subset are entirely independent of the group cardinality $|G|$, and therefore remain valid even when $G$ is infinite. The key insight is that random group elements are sufficient, with high probability, to enforce all necessary invariance constraints. This stands in sharp contrast to approaches based on generating sets, which either do not extend to infinite groups or require preprocessing the entire group, rendering them computationally infeasible.

\begin{algorithm}[t]
\caption{Randomized Subset Selection}
\label{alg}
\begin{algorithmic}[1] 
    \REQUIRE Query access to $D(g) \in \mathbb{R}^{r \times r}$, $\forall g \in G$;  number of iterations $T = \mathcal O\!\left(r^2+r\log\frac{1}{\delta}\right)$
    \ENSURE A subset $S \subseteq G$

    \STATE \textbf{Initialization:} $S \gets \{\mathrm{id}_G\}$ and $\mathcal{B}_S \gets \{e_i : i \in [r]\} \subseteq \mathbb{R}^r$, where $e_i$ denotes the $i$-th standard basis vector.
    \FOR{$t = 1, \dots, T$}
        \STATE Sample $g \sim \mathrm{Unif}(G)$
        \STATE Sample 
        $
        x = \sum_{v \in \mathcal{B}_S} N_v v,
        $
        where $N_v \sim \mathcal{N}\!\left(0, \tfrac{1}{|\mathcal{B}_S|}\right)$ independently for all $v \in \mathcal{B}_S$
        \IF{$\| (D(g) - I_r) x \|_2^2 > 0$}
            \STATE $S \gets S \cup \{ g \}$
            \STATE $\mathcal{B}_S \gets$ an orthonormal basis of 
            $
            \operatorname{span}(\mathcal{B}_S) \cap \ker(I_r - D(g))
            $
        \ENDIF
    \ENDFOR
    \STATE \textbf{return} $S$
\end{algorithmic}
\end{algorithm}

\begin{proposition}\label{prop:algo}
For each $g \in G$, define $V_g \coloneqq \ker(I_r - D(g))$. If $T = \mathcal O\!\left(r^2+r\log\frac{1}{\delta}\right)$,
then with probability at least $1-\delta$, Algorithm~\ref{alg} returns a subset
$S \subseteq G$ with $|S| \le r$
such that
\[
\bigcap_{g \in S} V_g \;=\; \bigcap_{g \in G} V_g.
\]
\end{proposition}

With the randomized subset selection procedure in place, we are now ready to state our main algorithmic and statistical guarantee. The resulting method builds upon the spectral-averaging framework of \citet{soleymanilearning}, while replacing generating sets with random subsets.

\begin{theorem}\label{thm}
Let $G$ be a group, and let $\{(x_i,y_i)\}_{i=1}^n$ be a labeled dataset sampled from a $d$-dimensional manifold $\mathcal{M}$. Suppose the target regression function satisfies
$f^\star \in \mathcal{H}^s(\mathcal{M})$ for some $s > d/2$, and define $\alpha \coloneqq 2s/d$ (Sobolev regression), or alternatively that $\mathcal{H}$ is finite-dimensional.

\noindent
Then, \Cref{alg:all}, achieved via spectral averaging with the randomized subset selection of \Cref{alg}, running in $\operatorname{poly}(n,d,\log(1/\delta))$ time, returns, with probability at least $1-\delta$, an exactly $G$-invariant estimator $\widehat{f}$. The estimator achieves excess population risk
$\mathcal{R}(\widehat{f})
=
\mathcal{O}\!\left(n^{-s/(s+d/2)}\right)$ for Sobolev regression,
or
$\mathcal{O}(\dim(\mathcal{H}^G)/n)$
in finite-dimensional settings. Thus, in the finite-dimensional case, this rate is minimax optimal over the class of $G$-invariant functions in $\mathcal{H}$.
\end{theorem}

The above theorem establishes that exact invariance can be enforced in polynomial time without prior knowledge of the group structure, while simultaneously achieving statistically optimal sample complexity in finite-dimensional settings.

\paragraph{Proof sketch for Proposition~\ref{prop:algo}}

A central technical contribution of the paper is Proposition~\ref{prop:algo}, which enables learning with exact invariances for arbitrary groups. The proposition shows that by randomly sampling group elements, one can construct a subset $S \subseteq G$, whose size is independent of $|G|$, such that
\[
\bigcap_{g \in S} V_g \;=\; \bigcap_{g \in G} V_g .
\]

At first glance, this may seem surprising. The space $\bigcap_{g \in G} V_g$ corresponds exactly to the subspace of invariant functions, whereas $\bigcap_{g \in S} V_g$ consists of functions invariant under the subgroup generated by $S$. While any generating set $S$ of $G$ trivially satisfies this equality, a key observation is that generation of the group is not necessary: since we only observe the action of $G$ through its representation on $\mathbb{R}^r$, distinct group elements may induce redundant linear constraints.

The proof exploits this representation-theoretic viewpoint. Starting from the full space $\mathbb{R}^r$, each sampled group element $g$ induces a constraint $D(g)f=f$, shrinking the feasible subspace to $V_g=\ker(I_r-D(g))$. Whenever the intersection $\bigcap_{g \in S} V_g$ has dimension strictly larger than that of $\bigcap_{g \in G} V_g$, a randomly sampled group element has a nontrivial probability of further reducing this dimension. Algorithm~\ref{alg} implements this idea by iteratively sampling group elements and updating the intersection whenever a new constraint is informative.

Since the dimension of the invariant subspace is at most $r$, and each successful iteration reduces the dimension by at least one, after at most $r$ effective reductions, occurring with high probability, the algorithm identifies a subset $S$ for which
\[
\bigcap_{g \in S} V_g = \bigcap_{g \in G} V_g .
\]
Importantly, the total number of iterations required depends only on $r$, and not on the size or structure of the group $G$. This establishes the main technical ingredient behind our first contribution.

\begin{remark}
When $G$ is the trivial group, Algorithm~\ref{alg:all}  reduces to the standard projection estimator in the chosen feature space. In particular, under the uniform distribution and an orthonormal feature basis, the population feature covariance is the identity, so the estimator can be viewed as the ordinary least-squares/projection estimator with the covariance plug-in replaced by its population value. Thus, in this special case, the algorithm agrees with ordinary linear estimation in the feature space up to the usual distinction between empirical OLS and projection using the identity covariance.
\end{remark}

\begin{algorithm}[t]
\caption{Spectral Averaging with Randomized Subset Selection}
\label{alg:all}
\begin{algorithmic}[1]
\REQUIRE Dataset $\mathcal{S} = \{(x_i, y_i)\}_{i=1}^n$ and $\alpha = 2s/d > 1$ for Sobolev regression
\ENSURE $G$-invariant estimator $\widehat{f}$

\STATE Set $r \gets n^{1/(1+\alpha)}$ for Sobolev regression; otherwise set $r \gets \dim(\mathcal{H})$
\FOR{each $\ell \in [r]$}
    \STATE $\widetilde{f}_{\ell} \gets \frac{1}{n} \sum_{i=1}^n y_i \phi_{\ell}(x_i)$
\ENDFOR
\STATE $S \gets$ Randomized Subset Selection for $G$ and $\mathcal{H}$ \hspace{6em}(\Cref{alg})
\STATE Solve the linearly constrained quadratic program:
\[
\widehat{f}
\gets
\argmin_{f \in \mathbb{R}^r}
\sum_{\ell=1}^{r} (f_{\ell} - \widetilde{f}_{\ell})^2,
\qquad
\text{s.t.}\quad D(g)f = f \;\; \forall g \in S
\]
\STATE \textbf{Return} $\widehat{f}(x) = \sum_{\ell=1}^{r} \widehat{f}_{\ell} \phi_{\ell}(x)$
\end{algorithmic}
\end{algorithm}

\subsection{Sobolev Rates and Computational--Statistical Trade-offs}

In the Sobolev setting, our guarantees should be interpreted relative to the
computational cost of enforcing invariance. When $G$ is trivial, our rate reduces
to the classical minimax-optimal Sobolev regression rate. For nontrivial group
actions, sharper invariant rates are known information-theoretically
\cite[Theorem~4.1]{tahmasebi2023exact}; however, achieving these rates relies on
a group-adapted regularization scheme \cite[Eq.~122]{tahmasebi2023exact} and, in
its kernel implementation, on averaging over the entire group. In our setting,
this corresponds to a potentially large spectral truncation and a computational
procedure that is infeasible when full group averaging is expensive.

Our contribution is different and complementary: we give a polynomial-time
procedure that enforces exact invariance while attaining the full convergence
rate of the corresponding no-invariance Sobolev problem. Thus, even though our
Sobolev guarantee is not claimed to be the sharp invariant minimax rate, it
achieves a statistically meaningful rate under exact invariance without requiring
full group averaging. Determining the optimal computational--statistical
trade-off for Sobolev spaces with invariances remains open. Resolving this
question likely requires a finer understanding of how the group action interacts
with low-frequency Laplace--Beltrami eigenspaces. In particular, it is currently
unclear whether the sharp invariant minimax rate can be achieved in polynomial
time without full group averaging, or whether an inherent computational gap is
present. In contrast, in the finite-dimensional setting, our method achieves the
optimal sample complexity of learning under invariances in polynomial time.

 \subsection{Symmetry Discovery for Learning with Exact Invariances}

We now extend the previous setting to the case where the invariance group is \emph{unknown}. Specifically, we assume that there exists an unknown subgroup
$H \subseteq G$ such that the target function $f^\star \in \mathcal{H}$ is invariant under $H$ and under no larger subgroup. Here, $G$ denotes a known \emph{reference finite group}. 

More precisely, to ensure that information about the subgroup $H$ is sufficiently present in the target function, we assume that $f^\star$ is drawn \emph{isotropically} from the subspace of $H$-invariant functions within $\mathcal{H}$. In the Sobolev setting, isotropy is defined with respect to the truncated function class of dimension $r \coloneqq \mathcal{O}(\sqrt{n})$.
 This assumption rules out pathological target functions for which the underlying symmetry would be statistically undetectable. Moreover, to avoid identifiability issues, we identify any two subgroups that induce the same invariant function class.

A key observation is that, once a generating set $S \subseteq G$ for $H$ is identified, learning reduces immediately to the exact-invariance setting studied in the previous section by simply running Algorithm~\ref{alg:all} with this set of constraints. The central challenge in symmetry discovery is therefore to identify a generating set for $H$ using only labeled data.

To make this problem tractable, we assume that $H$ has bounded index in $G$, namely that
\[
|H| \ge \kappa |G|
\quad \text{for some } \kappa \in (0,1].
\]
Under this assumption, a uniformly random element $g \sim \mathrm{Unif}(G)$ lies in $H$ with probability $\kappa$. Consequently, by sampling
\[
N = \Omega\!\left(\frac{\log |G| + \log(1/\delta)}{\kappa}\right)
\]
independent elements from $G$, we obtain, with probability at least $1-\delta$, a subset of size $\mathcal{O}(\log |H|)$ contained in $H$. Classical results on random Cayley graphs \citep{alon1994random} imply that such a random subset generates $H$ with high probability.

The remaining question is therefore algorithmic: given a fixed $g \in G$, can we test from data whether $f^\star$ is invariant under $g$, i.e.,
whether $f^\star(gx) = f^\star(x)$ holds almost surely?

Our approach answers this question via a data-driven hypothesis test. For each candidate element $g \in G$, we compare two estimators obtained via spectral averaging: one unconstrained estimator, and one estimator constrained to be invariant under $g$. Using a held-out test set, we evaluate whether imposing the constraint $D(g)f=f$ reduces the empirical test error. Elements for which this constraint leads to improved generalization performance are retained. Collecting such elements yields a candidate generating set for $H$.

This procedure leads to the following guarantee for symmetry discovery.

\begin{theorem}\label{thm:disc}
Let $G$ be a finite reference group, and consider the family of subgroups
$H \le G$ satisfying $|H| \ge \kappa |G|$ for some $\kappa \in (0,1]$.
Let $\{(x_i,y_i)\}_{i=1}^n$ be a labeled dataset sampled from a $d$-dimensional manifold $\mathcal{M}$.
Assume the unknown $H$-invariant target function satisfies
$f^\star \in \mathcal{H}^s(\mathcal{M})$ for some $s > d/2$, and define $\alpha \coloneqq 2s/d$ (Sobolev regression), or alternatively that $\mathcal{H}$ is finite-dimensional. Moreover, $f^\star \in \mathcal{H}$ is drawn \emph{isotropically}, according to the definition above.

\noindent
Then Algorithm~\ref{alg:discovery} returns, with probability at least $1-\delta$, an exactly $H$-invariant estimator $\widehat{f}$ together with a generating set for $H$, in
$\operatorname{poly} \Big(n,d,\tfrac{1}{\kappa},\log |G|,\log\tfrac{1}{\delta}\Big)$
time. The estimator achieves excess population risk
$\mathcal{R}(\widehat{f})
=
\mathcal{O}\!\left(n^{-s/(s+d/2)}\right)$ for Sobolev regression,
or
$\mathcal{O}(\dim(\mathcal{H}^G)/n)$
in finite-dimensional settings. Thus, in the finite-dimensional case, it  achieves the minimax optimal rate over the class of $H$-invariant functions in $\mathcal{H}$.
\end{theorem}

Theorem~\ref{thm:disc} shows that even when the symmetry group is unknown, exact invariances can be discovered and enforced in polynomial time while retaining statistically optimal rates. This establishes a sharp statistical–computational trade-off for learning under exact invariances and provides theoretical justification for the empirical success of symmetry discovery methods.

\begin{algorithm}[t]
\caption{Symmetry Discovery via Spectral Averaging}
\label{alg:discovery}
\begin{algorithmic}[1]
\REQUIRE Training set $\mathcal{S}=\{(x_i,y_i)\}_{i=1}^n$; test set $\mathcal{S}'=\{(x'_i,y'_i)\}_{i=1}^n$; parameter $\alpha=2s/d>1$ for Sobolev regression; number of trials $T$; threshold $\eta$
\ENSURE $H$-invariant estimator $\widehat{f}$ and a generating set $S$ for $H$

\STATE Set $r \gets n^{1/(1+\alpha)}$ for Sobolev regression; otherwise set $r \gets \dim(\mathcal{H})$
\FOR{each $\ell \in [r]$}
    \STATE $\widetilde{f}_{\ell} \gets \frac{1}{n} \sum_{i=1}^n y_i \phi_{\ell}(x_i)$
\ENDFOR

\STATE Compute the unconstrained estimator:
$
\widehat{f}_{\emptyset}
\gets \widetilde{f} \in \mathbb{R}^r
$

\STATE Initialize $S \gets \emptyset$
\FOR{$t = 1,\ldots,T$}
    \STATE Sample $g \sim \mathrm{Unif}(G)$
    \STATE Compute the $g$-invariant estimator:
    $
    \widehat{f}_g
    \gets
    \argmin\limits_{f \in \mathbb{R}^r}
    \sum_{\ell=1}^{r} (f_{\ell} - \widetilde{f}_{\ell})^2
    \quad \text{s.t.} \quad D(g)f = f
    $
\STATE \textbf{If}
$
    \frac{1}{n}\sum_{i=1}^{n}
    \bigl(\widehat f_g(x_i')-y_i'\bigr)^2
    \le
    \frac{1}{n} \sum_{i=1}^{n}
    \bigl(\widehat f_{\emptyset}(x_i')-y_i'\bigr)^2
    + \eta
$  \textbf{then} $S\gets S\cup\{g\}$.
\ENDFOR

\STATE Compute the final estimator:
$
\widehat{f}
\gets
\argmin_{f \in \mathbb{R}^r}
\sum_{\ell=1}^{r} (f_{\ell} - \widetilde{f}_{\ell})^2
\quad \text{s.t.}\quad D(g)f = f \;\; \forall g \in S
$

\STATE \textbf{Return} $\widehat{f}(x)=\sum_{\ell=1}^{r} \widehat{f}_{\ell}\phi_{\ell}(x)$
\end{algorithmic}
\end{algorithm}

\paragraph{Scope of the results.}
Our symmetry-discovery guarantee is stated for finite groups. The present
approach relies on finite-group tools, including random Cayley graph
constructions, and therefore does not directly extend to infinite groups. An
extension to compact Lie groups would likely require different techniques,
possibly exploiting the local and representation-theoretic structure of Lie
groups; we leave this direction for future work.

The isotropy assumption on $f^\star$ is also necessary for discovery. Without a
separation condition, the target function may be invariant, or nearly invariant,
under several distinct subgroups, making it impossible to identify a unique
underlying group from data. Our assumption rules out this ambiguity and ensures
that the relevant group is statistically identifiable.

Importantly, this assumption is needed only for symmetry discovery, not for
invariant regression itself. Once a candidate subgroup $H$ is specified, even if
several subgroups are compatible with the data, Algorithm~\ref{alg:all} can be applied to
enforce $H$-invariance and obtain the convergence guarantee of
Theorem~\ref{thm} for that subgroup.

\section{Conclusion}\label{sec:conclusion}

We presented the first \emph{randomized} polynomial-time algorithm for learning with \emph{exact invariances} under general group actions, accommodating both finite and infinite groups. This result represents a substantial step toward understanding the computational complexity of learning with symmetries, and clarifies the algorithmic boundary between tractable and intractable instances of invariant learning. For finite groups, prior work showed that spectral averaging combined with group generators yields a deterministic polynomial-time algorithm \citep{soleymanilearning}. In contrast, for infinite groups no such deterministic procedure is currently known. Our results demonstrate that randomization suffices to overcome this barrier: spectral averaging with randomized subset selection enforces exact invariance in polynomial time without any dependence on the group cardinality. Whether a fully \emph{deterministic} polynomial-time algorithm exists for learning with exact invariances under infinite groups remains an intriguing open question.

We further addressed the problem of \emph{symmetry discovery}, where the invariance group is unknown. Under a bounded-index assumption, we showed that a simple sampling strategy, grounded in the theory of random Cayley graphs, enables efficient identification of the underlying symmetry group directly from data. Combining symmetry discovery with invariant learning yields estimators that are both computationally efficient and statistically optimal.

Taken together, our results provide the first rigorous statistical–computational guarantees for learning with exact invariances and for discovering unknown symmetries, offering a theoretical foundation for the growing empirical success of symmetry-aware learning methods.


\section*{Acknowledgments}
AS and PJ were partially supported by the ONR grant N00014-24-1-2470.
BT and SJ were partially supported by the Office of Naval Research under grant N00014-20-1-2023
(MURI ML-SCOPE) and by the National Science Foundation under award CCF-2112665
(TILOS AI Institute). SJ also acknowledges support from an Alexander von Humboldt Professorship.
 
\bibliography{references}
\bibliographystyle{plainnat}

\clearpage
\appendix

\section{Preliminaries on Groups and Manifolds}
\label{app:group-preliminaries}

We collect here the basic notions from group actions and Riemannian geometry
used throughout the paper. Our data domain is a smooth, compact, boundaryless
manifold $\mathcal M$. This assumption is made for clarity of presentation and
to avoid technicalities related to boundary conditions, noncompactness, or
singularities. The same ideas extend to other sufficiently regular data domains,
provided that the corresponding function spaces, measures, and group actions
are well defined.

\paragraph{Groups.}
A group is a set $G$ equipped with a binary operation $(g,h)\mapsto gh$
satisfying the following three axioms: (I) associativity, $(gh)k=g(hk)$ for all
$g,h,k\in G$; (II) existence of an identity element $e\in G$ such that
$eg=ge=g$ for all $g\in G$; and (III) existence of inverses, meaning that for
every $g\in G$ there exists $g^{-1}\in G$ with
$gg^{-1}=g^{-1}g=e$. Standard examples include finite cyclic groups,
permutation groups such as the symmetric group $S_d$, finite groups of
rotations, matrix groups such as the orthogonal group $O(d)$ and the special
orthogonal group $SO(d)$, and compact Lie groups acting by smooth
transformations.

\paragraph{Group actions on manifolds.}
A left action of a group $G$ on $\mathcal M$ is a map
\[
    G\times \mathcal M \to \mathcal M,
    \qquad
    (g,x)\mapsto gx,
\]
such that $ex=x$ and $(gh)x=g(hx)$ for all $g,h\in G$ and
$x\in\mathcal M$. Throughout the paper, we assume that the action is smooth,
meaning that each map $x\mapsto gx$ is a smooth diffeomorphism of
$\mathcal M$, and, when $G$ is a Lie group, the joint map $(g,x)\mapsto gx$ is
smooth.

The kernel of the action is
\[
    \ker(G\curvearrowright \mathcal M)
    :=
    \{g\in G:\ gx=x \text{ for all } x\in\mathcal M\}.
\]
Elements in the kernel act trivially on all data points and hence have no
statistical or computational effect in our setting. Therefore, without loss of
generality, one may replace $G$ by the effective quotient
\[
    G_{\mathrm{eff}} := G/\ker(G\curvearrowright \mathcal M),
\]
which acts faithfully on $\mathcal M$. Thus, throughout the paper, we identify
$G$ with its effective action and assume that the action is faithful (i.e., no non-identity element acts trivially on all of $\mathcal M$).

\paragraph{Compactness and Haar measure.}
Our results are stated for compact groups $G$. More precisely, after quotienting
out the kernel of the action, we assume that the effective transformation group
acting on $\mathcal M$ is compact. This assumption covers the main examples of
interest, including finite groups, compact matrix groups, and compact Lie group
actions. Compactness gives a canonical probability measure on the group: the
normalized Haar measure, which is the unique probability measure
invariant under left and right multiplication. This measure is the natural
notion of the uniform distribution on $G$ and is used whenever we sample random
group elements.

\paragraph{Group representations.}
A representation of a group $G$ on a finite-dimensional vector space $V$ is a
map $D:G\to \mathrm{GL}(V)$ such that
\[
    D(gh)=D(g)D(h),
    \qquad
    D(e)=I .
\]
Thus, a representation realizes each abstract group element $g\in G$ as an
invertible linear operator $D(g)$ acting on $V$, in a way that preserves the
group law. If $V$ is equipped with an inner product and each $D(g)$ preserves
this inner product, then the representation is called orthogonal over
$\mathbb R$ and unitary over $\mathbb C$; equivalently,
\[
    D(g)^\top D(g)=I
    \quad \text{or} \quad
    D(g)^*D(g)=I,
    \qquad \forall g\in G .
\]
In this paper, such representations arise by restricting the action of $G$ on
functions to a finite-dimensional feature space $\mathcal H$: after choosing a
basis of $\mathcal H$, the operator $T_g f(x)=f(g^{-1}x)$ is represented by a
matrix $D(g)$, and the identity $T_{gh}=T_gT_h$ becomes
$D(gh)=D(g)D(h)$.

\paragraph{Invariant Riemannian metrics and isometric actions.}
The compactness assumption also allows us to regard the action as isometric
without loss of generality. Indeed, starting from any Riemannian metric
$\mathfrak g_0$ on $\mathcal M$, we may average it over the group:
\[
    \mathfrak g_x(u,v)
    :=
    \int_G
    \mathfrak g_{0,g^{-1}x}
    \bigl(Dg^{-1}_x u, Dg^{-1}_x v\bigr)
    \,d\mu_G(g).
\]
The resulting metric $\mathfrak g$ is smooth, positive definite, and
$G$-invariant. Consequently, every transformation $x\mapsto gx$ is an isometry
of $(\mathcal M,\mathfrak g)$. The corresponding Riemannian volume measure is
therefore also $G$-invariant. In this sense, assuming that the group acts
isometrically with respect to the canonical volume measure is not restrictive
for compact group actions; it can always be enforced by choosing an invariant
Riemannian metric.

Once such a metric is fixed, the isometry group
\[
    \operatorname{Iso}(\mathcal M)
\]
is the group of all diffeomorphisms of $\mathcal M$ preserving the metric. By
the Myers--Steenrod theorem, $\operatorname{Iso}(\mathcal M)$ is a Lie group,
and for compact $\mathcal M$ it is compact. Thus, after choosing a
$G$-invariant metric, the group $G$ may be viewed as a compact subgroup of
$\operatorname{Iso}(\mathcal M)$. This provides the geometric setting used
throughout the paper: a compact group acting smoothly, faithfully, and
isometrically on a compact Riemannian manifold, equipped with the corresponding
invariant volume measure.

\paragraph{Setting and notation for Sobolev-space regression.}
Let $(\mathcal M,\mathfrak g)$ be a smooth, compact, connected, boundaryless
$d$-dimensional Riemannian manifold, and let $G$ act smoothly and isometrically
on $\mathcal M$, so that $x\mapsto gx$ is an isometry for every $g\in G$. We
observe i.i.d.\ samples $\mathcal S=\{(x_i,y_i)\}_{i=1}^n$, where $x_i$ is drawn
uniformly from the Riemannian volume measure on $\mathcal M$ and
\[
    y_i=f^\star(x_i)+\varepsilon_i .
\]
Here $f^\star\in \mathcal H^s(\mathcal M)$ for some $s>d/2$, and the noise
variables $\varepsilon_i$ are independent, mean zero, and have variance
$\sigma^2$.

Let $\Delta_{\mathcal M}$ denote the Laplace--Beltrami operator. By spectral
theory, there exists an $L^2(\mathcal M)$-orthonormal basis
$\{\varphi_{\lambda,\ell}\}_{\lambda,\ell}$ of eigenfunctions, grouped by
eigenvalues
\[
    0=\lambda_0 < \lambda_1 \le \lambda_2 \le \cdots
\]
with multiplicities $m_\lambda$, such that every $f\in L^2(\mathcal M)$ admits
the expansion
\[
    f(x)=\sum_{\lambda}\sum_{\ell=1}^{m_\lambda}
    f_{\lambda,\ell}\,\varphi_{\lambda,\ell}(x).
\]
In these coordinates, the Sobolev space $\mathcal H^s(\mathcal M)$ is defined
spectrally as
\[
    \mathcal H^s(\mathcal M)
    \coloneqq
    \left\{
        f=\sum_{\lambda}\sum_{\ell=1}^{m_\lambda}
        f_{\lambda,\ell}\varphi_{\lambda,\ell}
        :
        \|f\|_{\mathcal H^s(\mathcal M)}^2
        \coloneqq
        \sum_{\lambda}\sum_{\ell=1}^{m_\lambda}
        D_\lambda^{\alpha} f_{\lambda,\ell}^2
        <\infty
    \right\},
\]
where $\alpha\coloneqq 2s/d$
where $\alpha\coloneqq 2s/d$ and $D_\lambda$ denotes the cumulative spectral
dimension up to eigenvalue $\lambda$, namely
\[
D_\lambda \coloneqq \sum_{\lambda'\le \lambda} m_{\lambda'} .
\]
For each eigenvalue $\lambda$, let
\[
    V_\lambda
    \coloneqq
    \operatorname{span}\{\varphi_{\lambda,\ell}\}_{\ell=1}^{m_\lambda}.
\]
Since the action is isometric, the induced operator
\[
    T_g f(x)\coloneqq f(g^{-1}x)
\]
is unitary on $L^2(\mathcal M)$ and commutes with $\Delta_{\mathcal M}$. Hence
each eigenspace $V_\lambda$ is $G$-invariant, and the action restricts to an
orthogonal representation
\[
    D_\lambda(g):G\to O(m_\lambda)
\]
on the coefficient space of $V_\lambda$. This commutativity and the resulting
blockwise orthogonal action are discussed in detail by
\citet{soleymanilearning}. Writing
\[
    f_\lambda\coloneqq (f_{\lambda,\ell})_{\ell=1}^{m_\lambda},
\]
the invariance condition on the $\lambda$-th eigenspace is
\[
    D_\lambda(g) f_\lambda = f_\lambda
    \qquad \text{for all } g\in G .
\]

\paragraph{Invariant subspaces in finite-dimensional feature spaces.}
Let $\mathcal H\subseteq L^2(\mathcal M)$ be a finite-dimensional Hilbert
feature space that is closed under the action of $G$. The induced action on
functions is given by
\[
    T_g f(x) \coloneqq f(g^{-1}x),
    \qquad g\in G,
\]
so that $T_g\mathcal H\subseteq \mathcal H$. After choosing a basis of
$\mathcal H$ with $\dim(\mathcal H)=r$, each $T_g$ is represented by a matrix
$D(g)\in\mathbb R^{r\times r}$. The subspace of $G$-invariant functions is
\[
    \mathcal H^G
    \coloneqq
    \{f\in\mathcal H:T_g f=f,\ \forall g\in G\}.
\]
Equivalently, defining
\[
    V_g \coloneqq \{f\in\mathcal H:T_g f=f\},
\]
we have
\[
    \mathcal H^G=\bigcap_{g\in G} V_g .
\]
In coordinates, this is the common fixed subspace
\[
    \mathcal H^G
    \equiv
    \bigcap_{g\in G}\ker(D(g)-I).
\]
Thus, in the finite-dimensional setting, enforcing invariance amounts to
representing this intersection of fixed spaces. A central goal of our analysis
is to identify small subsets $S\subseteq G$ for which
\[
    \bigcap_{g\in S} V_g
    =
    \bigcap_{g\in G} V_g,
\]
or for which the corresponding approximation is sufficient for the desired
statistical guarantees.
\section{Proofs}

\subsection{Proof of Proposition~\ref{prop:algo}} \label{app:proof_prop} 
For any subset $S \subseteq G$, let us define a probability measure $\mu_S$ as follows.  
Consider the set $\mathcal{B}_S$, which is an orthonormal basis for the subspace $\bigcap_{g \in S} V_g$.  
We construct $\mu_S$ as the distribution of random linear combinations of vectors in this basis:
\[
\mu_S :=~\text{law of}~\sum_{v \in \mathcal{B}_S} N_v v,
\]
where the coefficients $N_v$ are independent Gaussian random variables, each drawn as $N_v \sim \mathcal{N}(0, \tfrac{1}{|\mathcal{B}_S|})$.  
This choice ensures that $\mu_S$ is isotropic in the span of $\mathcal{B}_S$, i.e., the covariance of the distribution is proportional to the identity restricted to $\operatorname{span}(\mathcal{B}_S)$.  
Intuitively, this means that $\mu_S$ places uniform weights  along all directions in the subspace $\bigcap_{g \in S} V_g$.

Now define the functional
\begin{align}
    A(S) := \mathbb{E}_{g} \, \mathbb{E}_{x \sim \mu_S} \| (D(g) - I_r)x\|_2^2,
\end{align}
where $g \in G$ is sampled uniformly at random.  
This quantity measures, in expectation, how much the action of $D(g)$ deviates from the identity transformation on vectors $x$ sampled from $\mu_S$.  
In other words, $A(S)$ introduces some kind of discrepancy between invariance under the full group $G$ and invariance under the restricted subset $S$.

\medskip\noindent\textbf{Step 1.} The case $A(S) = 0$. 
If $A(S) = 0$, then for every $g \in G$ and every $x$ in the support of $\mu_S$ we must have
\[
(D(g) - I_r)x = 0,
\]
i.e., $D(g)x = x$. This means that $x \in \bigcap_{g \in G} V_g$.

Since the support of $\mu_S$ is exactly $\operatorname{span}(\mathcal{B}_S)$, it follows that the entire subspace $\operatorname{span}(\mathcal{B}_S)$ is fixed by the group $G$.  We claim that $\operatorname{span}(\mathcal{B}_S) = \bigcap_{g \in S} V_g$, which follows from the way we update it in Algorithm \ref{alg}. 
This implies
\[
\bigcap_{g \in S} V_g \subseteq \bigcap_{g \in G} V_g \implies 
\bigcap_{g \in S} V_g =\bigcap_{g \in G} V_g,
\]
so in this case the invariant subspace with respect to $G$ is already fully captured by the smaller intersection over $S$.

\medskip\noindent\textbf{Step 2.} The case $A(S) > 0$. 
Suppose now that $A(S) > 0$. We will prove a quantitative lower bound on $A(S)$. Namely, we prove  that
\[
A(S) \ge \frac{2}{r}.
\]
Expanding the square inside the definition of $A(S)$ gives
\begin{align}
    A(S) &= \mathbb{E}_{g} \mathbb{E}_{x \sim \mu_S} 
            \Big[\|x\|_2^2 + \|D(g)x\|_2^2 -  x^\top D(g)x - x^\top D(g)^\top x \Big].
\end{align}
Since the representation $D(g)$ is orthogonal, we have $\|D(g)x\|_2^2 = \|x\|_2^2$.  
This allows us to rewrite the expression as:
\begin{align}
    A(S) &= \mathbb{E}_{g} \mathbb{E}_{x \sim \mu_S}
            \big[\|x\|_2^2 + \|x\|_2^2 - x^\top D(g)x - x^\top D(g)^\top x\big] \\
          &= 2 - \mathbb{E}_{g} \mathbb{E}_{x \sim \mu_S}
                            \big[ x^\top D(g) x + x^\top D(g)^\top x\big],
\end{align}
where in above we used the fact that $\mathbb{E}_{x \sim \mu_S}[\|x\|_2^2] = 1$, according to the definition of $\mu_S$.

But since $D(g)$ is orthogonal, averaging $D(g)$ or $D(g)^\top$ over $g \in G$ yields the same expectation.  
We therefore obtain
\[
A(S) = 2 - 2\mathbb{E}_{x \sim \mu_S}[x^\top \bar{D} x],
\]
where we define the average 
\[
\bar{D} := \mathbb{E}_g[D(g)] = \mathbb{E}_g[D(g)^\top].
\]

\medskip\noindent\textbf{Step 3.} Structure of $\bar{D}$.  
From basic representation theory, the matrix $\bar{D}$ is the orthogonal projection onto the subspace of invariant vectors (Schur's lemma).  
Equivalently, there exists an orthogonal change of basis $U$ such that
\[
\bar{D} = U P U^\dagger,
\]
where $P$ is the projection matrix onto the first $r_{\operatorname{inv}}$ coordinates, with $r_{\operatorname{inv}} = \dim\left (\bigcap_{g \in G} V_g\right)$.  In other words, 
\[
\bar{D} = U \;
\begin{bmatrix}
I_{r_{\operatorname{inv}}} & 0 \\
0 & 0
\end{bmatrix} U^\dagger.
\]
Thus, $\bar{D}$ corresponds to selecting exactly those elements invariant under the full group.

Moreover, since $S \subseteq G$, we automatically have
\[
\bigcap_{g \in G} V_g \subseteq \bigcap_{g \in S} V_g.
\]
That is, the invariant subspace for the full group is always contained within the invariant subspace defined by any subset of group elements.  
Therefore, $\operatorname{span}(\mathcal{B}_S)$ contains the true invariant subspace, and the expectation $\mathbb{E}_{x \sim \mu_S}[x^\top \bar{D} x]$ reflects the fraction of $L^2$-norm of the elements in the support of $\mu_S$ lying in this smaller subspace.

\medskip\noindent\textbf{Step 4.} Dimension ratio interpretation.  
Since $\mu_S$ is isotropic over $\bigcap_{g \in S} V_g$, the expectation of the quadratic form $x^\top \bar{D} x$ is equal to the ratio
\[
\frac{\dim(\bigcap_{g \in G} V_g)}{\dim(\bigcap_{g \in S} V_g)}.
\]
Substituting this back, we find
\begin{align}
    A(S) &= 2\Big(1- \frac{\dim(\bigcap_{g \in G} V_g)}{\dim(\bigcap_{g \in S} V_g)}\Big).
\end{align}

\medskip\noindent\textbf{Step 5.} Lower bound on $A(S)$. 
If $A(S) > 0$, then necessarily
\[
\dim(\bigcap_{g \in S} V_g) > \dim(\bigcap_{g \in G} V_g).
\]
The smallest possible difference between the two dimensions is exactly one, so
\[
\frac{\dim(\bigcap_{g \in G} V_g)}{\dim(\bigcap_{g \in S} V_g)} \le \frac{r-1}{r}.
\]
Therefore,
\begin{align}
    A(S) &\ge 2\Big(1- \frac{r-1}{r}\Big) = \frac{2}{r}.
\end{align}
This establishes the claimed lower bound.

\medskip\noindent\textbf{Step 6.} Probabilistic argument.
It remains to control the number of random trials needed to detect all strict
dimension decreases. By the previous step, whenever a strict discrepancy remains,
namely
\[
    \bigcap_{g\in G} V_g
    \subsetneq
    \bigcap_{g\in S} V_g,
\]
a fresh random draw succeeds in detecting it with probability at least $2/r$.
Indeed, if $X\in[0,1]$ denotes the corresponding detection statistic, then
$\mathbb E X\ge 2/r$, and hence
\[
    \mathbb P(X>0)\ge \mathbb E X \ge \frac{2}{r}.
\]
Let $Z$ be the number of successful detections among $T$ independent trials.
Then $Z$ stochastically dominates a binomial random variable
$\operatorname{Bin}(T,2/r)$. Therefore, by a standard Chernoff bound,
\[
    \mathbb P(Z<r)\le \delta
\]
provided that
\[
    T \ge C r\left(r+\log\frac{1}{\delta}\right)
\]
for a universal constant $C>0$. Since the dimension can decrease at most $r$
times, $r$ successful detections suffice to certify that no further strict
discrepancy remains. Thus, with probability at least $1-\delta$, the procedure
terminates succesfully  after
\[
    T = \mathcal O\!\left(r^2+r\log\frac{1}{\delta}\right)
\]
independent trials.

\medskip\noindent\textbf{Conclusion.}  
Putting everything together, with $T$ iterations we guarantee that, with probability at least $1-\delta$, the subspace defined by $S$ coincides with the true invariant subspace:
\[
\bigcap_{g \in G} V_g = \bigcap_{g \in S} V_g,
\]
while $|S| \le r$ since at each iteration at most one group element is chosen.  This completes the proof.

\subsection{Proof of Theorem \ref{thm}}

The proof follows the structure of the finite-group proof \citep[Theorem~1]{soleymanilearning} and differs
only in how the constraint set $S$ is chosen and analyzed; the statistical analysis is mostly unchanged.

\medskip\noindent\textbf{Step 1: Spectral reduction and decoupled convex programs.}
Because $\Delta_\mathcal{M}$ commutes with all isometries (hence with the $G$-action),
$G$ preserves each eigenspace $V_\lambda$ and acts on it through an orthogonal matrix $D_\lambda(g)$.
Therefore $f$ is $G$-invariant if and only if $D_\lambda(g)f_\lambda=f_\lambda$ for all $\lambda$ and all $g\in G$.
As in the finite-group analysis, minimizing the population risk $\mathbb{E}\|f-f^\star\|_{L^2}^2$ over
$G$-invariant $f$ reduces to \emph{independent} quadratic problems (QP) on the retained eigenspaces
$V_\lambda$ with linear constraints $D_\lambda(g)u=u$.
Replacing the intractable population coefficients by their empirical means $\widehat f_{\lambda,\ell}$ yields
the \emph{empirical} QPs mentioned above; their minimizers are the orthogonal projections of
$\widehat f_\lambda$ onto the fixed-point subspaces~\citep{soleymanilearning}.

\medskip\noindent\textbf{Step 2: A single random subset $S$ suffices for \emph{all} retained eigenspaces.}
Define the block-diagonal representation
\[
R(g)\;:=\;\bigoplus_{\lambda:\,D_\lambda\le D} D_\lambda(g)\in O(r),
\qquad r:=\sum_{\lambda:\,D_\lambda\le D}m_\lambda=D.
\]
Let $V_g:=\ker(I_r-R(g))$ be its fixed-point subspace.  Run \Cref{alg} (Randomized Subset Selection) \emph{once}
on the representation $R(\cdot)$ with $T=\Theta(r^2 + r\log(1/\delta))$ iterations to obtain a set $S\subseteq G$ of size
$|S|\le r$ such that, with probability at least $1-\delta$,
\[
\bigcap_{g\in S} V_g \;=\;\bigcap_{g\in G}V_g.
\]
This result follows from Proposition~\ref{prop:algo} and its proof is discussed in \Cref{app:proof_prop}:
the statistic
$A(S):=\mathbb{E}_{g}\,\mathbb{E}_{x\sim\mu_S}\| (R(g)-I_r)x\|_2^2$ either vanishes (in which case the intersections
coincide) or is bounded below by a positive constant depending on $r$; a standard concentration
argument then shows that each time $A(S)>0$ one detects it in $r$ trials and reduces the
candidate basis dimension by~$1$, hence $\mathcal{O}(r^2+r\log(1/\delta))$ trials suffice. 

Because $R(g)$ is block-diagonal, $V_g=\bigoplus_{\lambda:\,D_\lambda\le D}
\ker \bigl(I_{m_\lambda}-D_\lambda(g)\bigr)$ and therefore
\[
\bigcap_{g\in S}V_g=\bigoplus_{\lambda:\,D_\lambda\le D}\bigcap_{g\in S}\ker \bigl(I_{m_\lambda}-D_\lambda(g)\bigr),
\qquad
\bigcap_{g\in G}V_g=\bigoplus_{\lambda:\,D_\lambda\le D}\bigcap_{g\in G}\ker \bigl(I_{m_\lambda}-D_\lambda(g)\bigr).
\]
Thus the equality of intersections at the block level implies, \emph{for every retained}
$\lambda$, that
\[
\bigcap_{g\in S}\ker \bigl(I_{m_\lambda}-D_\lambda(g)\bigr)
\;=\;
\bigcap_{g\in G}\ker \bigl(I_{m_\lambda}-D_\lambda(g)\bigr).
\]
Consequently, projecting $\widehat f_\lambda$ onto the fixed-point subspace defined by $S$ is
the same as projecting onto the $G$-invariant subspace of $V_\lambda$.  Hence the resulting estimator
$\widetilde f$ is \emph{exactly $G$-invariant} with probability at least $1-\delta$.

Finally, note that by construction $r=D$.  Since $\alpha>1$, we have
$D=n^{1/(1+\alpha)}\le n^{1/2}$, hence $r=\mathcal{O}(\sqrt{n})$, matching the choice of $r$ in Proposition~\ref{prop:algo}.

\medskip\noindent\textbf{Step 3: Risk bound (classic bias-variance analysis).}
Write $f^\star=f^\star_{\le D}+f^\star_{>D}$ for the orthogonal decomposition into the retained and
discarded spectral parts.  Exactly as the proof of \citet[Theorem 1]{soleymanilearning}, we decompose
\[
\mathbb{E}\mleft[\|\widetilde f-f^\star\|_{L^2}^2 \mright]
\;\le\;2\,\mathbb{E}\mleft[ \|\widetilde f-f^\star_{\le D}\|_{L^2}^2\mright] \,+\,2\,\|f^\star_{>D}\|_{L^2}^2.
\]
The \emph{bias term} obeys $\|f^\star_{>D}\|_{L^2}^2\le D^{-\alpha}\|f^\star\|_{H^s(M)}^2$ by $f^\star \in H^s(\mathcal{M})$ and the spectral
definition of the Sobolev norm. This is because,
\[
\|f^\star_{>D}\|_{L^2}^2 & = \sum_{\lambda: D_\lambda > D} \sum_{\ell=1}^{m_\lambda} (f^\star_{\lambda,\ell})^2 \\
    & = \sum_{\lambda: D_\lambda > D} \sum_{\ell=1}^{m_\lambda} D_\lambda^{-\alpha} D_\lambda^{\alpha}(f^\star_{\lambda,\ell})^2 \\
    & \leq D^{-\alpha} \sum_{\lambda: D_\lambda > D} \sum_{\ell=1}^{m_\lambda} D_\lambda^{\alpha} (f^\star_{\lambda,\ell})^2 \\
    & \leq D^{-\alpha} \sum_{\lambda} \sum_{\ell=1}^{m_\lambda} D_\lambda^{\alpha} (f^\star_{\lambda,\ell})^2 \\
    & = D^{-\alpha} \|f^\star\|_{H^s(\mathcal{M})}^2.
\]
In turn, we focus on the \emph{variance term},
\[
\mathbb{E}\mleft[\|\widehat f_{\le D}-f^\star_{\le D}\|_{L^2}^2 \mright]
=\sum_{D_\lambda\le D}\sum_{\ell=1}^{m_\lambda}
\mathbb{E}\mleft[ \bigl|\widehat f_{\lambda,\ell}-f^\star_{\lambda,\ell}\bigr|^2 \mright].
\]
By definition, we obtain
\begin{align}
    f^\star_{\lambda,\ell} = \mathbb{E}_x[f^\star(x)\phi_{\lambda, \ell}(x)] = \mathbb{E}_{x,y}[y\phi_{\lambda, \ell}(x)],
\end{align}
for every $\lambda,\ell$. In addition, $\widetilde{f}_{\lambda,\ell}$ denotes the empirical estimate derived from the data:
\begin{align}
    \widetilde{f}_{\lambda,\ell} = \frac{1}{n} \sum_{i=1}^n y_i \phi_{\lambda,\ell}(x_i).
\end{align}
Thus, we get
\[
\mathbb{E}[|\widetilde{f}_{\lambda,\ell} - f^\star_{\lambda,\ell}|^2] &= \frac{1}{n} \mathbb{E}\left[| y \phi_{\lambda,\ell}(x) - \mathbb{E}[y\phi_{\lambda,\ell}(x)] |^2\right] \\
    & = \frac{1}{n} \mathbb{E}\left[| \epsilon \phi_{\lambda,\ell}(x) + f^\star(x) \phi_{\lambda,\ell}(x) - \mathbb{E}[f^\star(x)\phi_{\lambda,\ell}(x)] |^2\right] \\
    & = \frac{1}{n} \left( \sigma^2 \mathbb{E}[\phi_{\lambda,\ell}^2] + \mathbb{E}\left[|f^\star(x) \phi_{\lambda,\ell}(x) - \mathbb{E}[f^\star(x)\phi_{\lambda,\ell}(x)] |^2\right] \right) \\
    & \leq \frac{1}{n} \left( \sigma^2 + \mathbb{E}[f^\star(x)^2 \phi_{\lambda,\ell}^2(x)] \right) \\
    & \leq \frac{1}{n} \left( \sigma^2 + \|f^\star\|^2_{L^{\infty}(\mathcal{M})} \right),
\]
since the $\varphi_{\lambda,\ell}$'s are orthonormal and $\widehat f_{\lambda,\ell}$ are empirical means. Summing over dimensions up to $D$, we obtain
\[
    \mathbb{E}[\| \widetilde{f} - f^\star_{\leq D}\|^2_{L^2(\mathcal{M})}] \leq \frac{D}{n} \left( \sigma^2 + \|f^\star\|^2_{L^{\infty}(\mathcal{M})} \right).
\]

Because $\widetilde f$ is the \emph{orthogonal projection} (in each $V_\lambda$) of $\widehat f$ onto a
linear subspace, the projection can only reduce squared error, so
\[
\mathbb{E} \mleft[ \|\widetilde f-f^\star_{\le D}\|_{L^2}^2 \mright] \le \mathbb{E} \mleft[ \|\widehat f_{\le D}-f^\star_{\le D}\|_{L^2}^2 \mright] \leq \frac{D}{n} \left( \sigma^2 + \|f^\star\|^2_{L^{\infty}(\mathcal{M})} \right).
\]
Putting the two parts (\emph{bias} and \emph{variance}) together and taking $D=n^{1/(1+\alpha)}$ yields
\[
\mathbb{E}\|\widetilde f-f^\star\|_{L^2}^2
\;\le\; \left( \sigma^2 + \|f^\star\|^2_{L^{\infty}(\mathcal{M})} \right) \,\frac{D}{n}+ \mleft( \|f^\star\|_{H^s(\mathcal{M})}^2 \mright) \,D^{-\alpha}
\;=\;\mathcal{O}\!\left(n^{-\frac{\alpha}{1+\alpha}}\right),
\]
exactly as for the finite group settings.  \emph{All the calculations of this step are borrowed verbatim
from the proof of  for finite groups in \citet{soleymanilearning}}.

\medskip\noindent\textbf{Step 4: Running-time bound.}
Computing the primary coefficients $\widehat f_{\lambda,\ell}$ for $D_\lambda\le D$ takes
$\mathcal{O}\bigl(nD\bigr)=\mathcal{O}\bigl(n^{\frac{2+\alpha}{1+\alpha}}\bigr)$ time.
Forming the constraint matrices $\{D_\lambda(g)\}_{g\in S}$ costs
$\mathcal{O}\bigl(|S|\sum_{D_\lambda\le D}m_\lambda^2\bigr)\le \mathcal{O}\bigl(|S|\,D^2\bigr)$ oracle calls,
because $\sum m_\lambda^2\le (\sum m_\lambda)^2=D^2$.
Solving the QPs by the closed form uses a pseudoinverse of a matrix of size
$(|S|m_\lambda)\times(|S|m_\lambda)$ and hence time $\mathcal{O}\bigl(|S|^3m_\lambda^3\bigr)$ per $\lambda$;
summing gives $\mathcal{O}\bigl(|S|^3\sum m_\lambda^3\bigr)\le \mathcal{O}\bigl(|S|^3 D^3\bigr)$.
By Step~2, with probability at least $1-\delta$ we have $|S|=\mathcal{O}(D)$,
and since $D=n^{1/(1+\alpha)}\le n^{1/2}$, this is polynomial in $n$ (and independent of $|G|$).
Thus the total running time is $\mathrm{poly}\bigl(n,d,\log(1/\delta)\bigr)$.

\medskip
Combining Steps~1--4 concludes the proof: the estimator $\widetilde f$ is exactly $G$-invariant
(with probability $\ge 1-\delta$), achieves the same excess-risk rate as in the finite-group case,
and the algorithm runs in time polynomial in $n,d,\log(1/\delta)$, \emph{independent of the
cardinality of~$G$}. 

\begin{remark}
    The proof for the finite-dimensional setting proceeds analogously (except that we do not need to cap the dimension and bound the tail),  since the basis is projected onto the quotient space. From this point onward, the minimax optimality argument follows standard lines.
\end{remark}

\subsection{Proof of Theorem~\ref{thm:disc}}

In this section, we present a proof of the main result of the paper on symmetry discovery.

We begin by noting that, in light of the proof of Theorem~\ref{thm}, it suffices to show that Algorithm~\ref{alg:discovery} returns a set $S \subseteq G$ that generates the true symmetry group $H$ with high probability. Once such a generating set is obtained, exact $H$-invariance of the final estimator follows immediately from the construction of the algorithm, and the stated statistical guarantees then follow from standard results on invariant regression.

\paragraph{Sampling a generating set for $H$.}
We first argue that sampling
\[
T = \Omega\!\left(\frac{1}{\kappa}\big(\log |G| + \log \tfrac{1}{\delta}\big)\right)
\]
elements uniformly at random from $G$ is sufficient to ensure that, with probability at least $1-\delta$, the sampled elements contain a generating set for $H$.

It is a classical result in the theory of random Cayley graphs that, for any finite group $H$, drawing
\[
T_H = \Omega\!\left(\log |H| + \log \tfrac{1}{\delta}\right)
\]
elements uniformly at random from $H$ produces a generating set for $H$ with probability at least $1-\delta$; see, e.g., \citep{alon1994random}.

Since $|H| \ge \kappa |G|$, each draw from $\mathrm{Unif}(G)$ lands in $H$ with probability at least $\kappa$. By standard concentration bounds for binomial random variables, after
\[
T = \Omega\!\left(\frac{1}{\kappa}\big(\log |H| + \log \tfrac{1}{\delta}\big)\right)
\]
samples, the number of draws falling inside $H$ is at least $T_H$ with probability at least $1-\delta$. Consequently, with high probability, the sampled elements contain a generating set for $H$.

\paragraph{Acceptance and rejection of sampled elements.}
Having ensured that a generating set for $H$ appears among the sampled group elements, it remains to show that Algorithm~\ref{alg:discovery} correctly distinguishes elements of $H$ from elements in $G\setminus H$, where the latter denotes transformations $g\in G$ for which $T_g f^\star\neq f^\star$ (assuming that $H$ denotes the largest subgroup under which $f^\star$ is invariant, to resolve any ambiguity arising from identifiability issues).

For each randomly sampled element $g\in G$, there are two types of errors to control:
\begin{itemize}
\item If $g\in H$, we must ensure that the algorithm accepts $g$ with high probability.
\item If $g\notin H$, we must ensure that the algorithm rejects $g$ with high probability.
\end{itemize}

The remainder of the proof is devoted to establishing these two guarantees by comparing the population and empirical risks of the unconstrained estimator and the estimator constrained to be invariant under $g$. We show that invariance constraints corresponding to elements of $H$ do not increase the population risk, while invariance constraints corresponding to elements outside $H$ introduce a detectable bias with high probability under the isotropy assumption on $f^\star$.

For a candidate element $g\in G$, let
\[
    W \coloneqq \mathcal H^H,
    \qquad
    W_g \coloneqq \mathcal H^{\langle H,g\rangle}
    =
    W\cap \mathcal H^{\langle g\rangle},
\]
and let $P_g$ denote the orthogonal projector from $W$ onto $W_g$. Since
$f^\star\in W$, imposing invariance under $g$ introduces the population-level
bias
\[
    b_g
    \coloneqq
    \|(I-P_g)f^\star\|_{L^2(\mathcal M)}^2 .
\]
If $g\in H$, then $W_g=W$ and hence $b_g=0$. If $g\notin H$, then, under the
identifiability convention that $H$ is the maximal invariance subgroup of
$f^\star$, we have $W_g\subsetneq W$, and therefore $b_g>0$.

The key point is that this separation holds uniformly over the finitely many
candidate elements inspected by the algorithm. Let
$\mathcal G_{\mathrm{test}}\subseteq G$ denote this candidate set, with
$|\mathcal G_{\mathrm{test}}|=T$. Conditional on
$\mathcal G_{\mathrm{test}}$, the subspaces
$\{W_g:g\in\mathcal G_{\mathrm{test}}\setminus H\}$ are fixed strict subspaces
of $W$. Under the isotropy assumption, we may write $f^\star=R u$, where
$u$ is uniformly distributed on the unit sphere in $W$ and
$R=\|f^\star\|_{L^2(\mathcal M)}$. If $d_H\coloneqq \dim(W)$, then for any fixed
strict subspace $U\subsetneq W$ and any $\tau\in(0,1)$,
\[
    \mathbb P\!\left(\operatorname{dist}(u,U)^2\le \tau\right)
    \le C\sqrt{d_H\tau},
\]
for a universal constant $C>0$; the worst case is codimension one. A union bound
over the at most $T$ tested elements gives
\[
    \mathbb P\!\left(
    \exists g\in \mathcal G_{\mathrm{test}}\setminus H:
    b_g \le R^2\tau
    \right)
    \le
    C T\sqrt{d_H\tau}.
\]
Thus, choosing
\[
    \tau
    =
    \frac{c\delta^2}{T^2 d_H}
\]
for a sufficiently small universal constant $c>0$, we obtain, with probability
at least $1-\delta$, the simultaneous separation
\[
    b_g=0
    \quad \text{for all } g\in H,
    \qquad
    b_g\ge \gamma
    \quad \text{for all } g\in \mathcal G_{\mathrm{test}}\setminus H,
\]
where
\[
    \gamma
    \coloneqq
    R^2\tau
    =
    \Omega\!\left(\frac{R^2\delta^2}{T^2 d_H}\right).
\]
In particular, when $T=\operatorname{poly}(d)$ and
$d_H\le r=\operatorname{poly}(d)$, this gap is inverse-polynomial in $d$.

It remains to ensure that the empirical comparisons resolve this population
gap. Using an independent validation sample, standard concentration bounds imply
that, for all candidates in $\mathcal G_{\mathrm{test}}$ simultaneously, the
empirical losses of the unconstrained and $g$-constrained estimators are within
$o(\gamma)$ of their corresponding population losses, provided the validation
size is polynomial in $r$, $1/\gamma$, and $\log(T/\delta)$. On this event, a
thresholded comparison with tolerance, say $\eta = \gamma/4$, is correct: elements
$g\in H$ introduce no bias and are accepted, whereas elements $g\notin H$
introduce bias at least $\gamma$ and are rejected. Consequently, with high
probability over the isotropic draw of $f^\star$, Algorithm~\ref{alg:discovery}
correctly distinguishes all sampled elements of the group $H$ from all sampled elements
outside $H$.

\end{document}

%% file: notation.tex
\def\[#1\]{\begin{align*}#1\end{align*}}

\NewDocumentCommand{\numberthis}{om}{%
  \IfNoValueTF{#1}{%
    \refstepcounter{equation}\tag{\theequation}%
  }{%
    \tag{#1}%
  }%
  \label{#2}%
}

\newcommand{\restatethm}[1]{%
    \begingroup
    \csname #1*\endcsname 
    \endgroup
}

\usepackage{mleftright}

\crefname{lemma}{Lemma}{Lemmas}
\crefalias{lemma}{theorem}
\crefname{assumption}{Assumption}{Assumptions}
\crefalias{assumption}{theorem}
\crefname{proposition}{Proposition}{Propositions}
\crefalias{proposition}{theorem}

\DeclareMathOperator*{\argmin}{arg\,min}

\definecolor{transparentgray}{rgb}{0.5,0.5,0.5} 